\documentclass{article}

\usepackage{arxiv}
\usepackage[numbers,sort&compress]{natbib}

\usepackage[utf8]{inputenc} 
\usepackage[T1]{fontenc}    
\usepackage{hyperref}       
\usepackage{url}            
\usepackage{booktabs}       
\usepackage{amsfonts}       
\usepackage{nicefrac}       
\usepackage{microtype}      
\usepackage{lipsum}
\usepackage{graphicx}
\usepackage{amsmath} 
\usepackage{algorithm}
\usepackage{caption}
\usepackage{pgfplots}
\usepackage{tikz}
\usetikzlibrary{positioning,calc,decorations.pathreplacing}

\usepackage{pgfplotstable}
\pgfplotsset{compat=1.18}
\usepackage[noend]{algpseudocode} 

\usepackage[utf8]{inputenc}
\DeclareUnicodeCharacter{2211}{\ensuremath{\sum}}
\DeclareUnicodeCharacter{2208}{\ensuremath{\in}}
\DeclareUnicodeCharacter{2013}{\textendash}

\usepackage{algpseudocode} 

\graphicspath{ {./images/} }

\title{Performance Analysis of Supervised Machine Learning Algorithms for Text Classification \\[0.5em]
\large Published in 2016 19th International Conference on Computer and Information Technology (ICCIT)}

\author{
Sadia Zaman Mishu \\
Department of Computer Science and Engineering \\
Rajshahi University of Engineering and Technology \\
Rajshahi 6204, Bangladesh \\
\texttt{sadia\_cse09@yahoo.com} \\
\And
S M Rafiuddin \\
Department of Computer Science and Engineering \\
Rajshahi University of Engineering and Technology \\
Rajshahi 6204, Bangladesh \\
\texttt{torifat.cs@gmail.com} \\
}

\begin{document}
\maketitle
\begin{abstract}
The demand for text classification is growing significantly in web searching, data mining, web ranking, recommendation systems, and so many other fields of information and technology. This paper illustrates the text classification process on different dataset using some standard supervised machine learning techniques. Text documents can be classified through various kinds of classifiers. Labeled text documents are used to classify the text in supervised classifications. This paper applies these classifiers on different kinds of labeled documents and measures the accuracy of the classifiers. An Artificial Neural Network (ANN) model using Back Propagation Network (BPN) is used with several other models to create an independent platform for labeled and supervised text classification process. An existing benchmark approach is used to analysis the performance of classification using labeled documents. Experimental analysis on real data reveals which model works well in terms of classification accuracy. 
\end{abstract}


\section{Introduction}
Text classification of labeled documents is becoming increasingly necessary due to the rapid growth of documents across the World Wide Web (WWW). Text classification is the task of assigning a document to a predefined category \citep{Ikonomakis2005}. In this research, several machine learning algorithms are applied to multiple datasets to determine the accuracy of the classifiers. The machine learning algorithms explored include Na{\"i}ve Bayes techniques such as Multinomial Na{\"i}ve Bayes and Bernoulli Na{\"i}ve Bayes; linear classifiers including Logistic Regression and Stochastic Gradient Descent; Support Vector Machine models such as Support Vector Clustering (SVC) and Linear SVC; and Artificial Neural Network models such as the Back Propagation Network. These algorithms are evaluated on experimental datasets such as the Reuters Corpus \citep{ReutersCorpus}, the Brown Corpus \citep{BrownCorpus}, and the Movie Review Corpus \citep{CornellReviews}.

Formally, let $D = \{d_1, d_2, \ldots, d_n\}$ be a set of documents, and let $C = \{c_1, c_2, \ldots, c_m\}$ be the set of classes. The text classification process aims to assign each document $d_i$ to the appropriate class $c_j$ \citep{Ikonomakis2005}. This research addresses hard categorization of text, where every document must be placed into exactly one of the specific classes.

The text classification process generally involves the following steps \citep{Ikonomakis2005}:
\begin{enumerate}
    \item Reading documents
    \item Tokenizing texts
    \item Stemming
    \item Stop word deletion
    \item Vector representation of text
    \item Feature selection
    \item Applying supervised learning algorithms
    \item Measuring accuracy
\end{enumerate}

Different metrics can be used to measure classification performance; in this research, the F1 score is employed as the primary evaluation metric \citep{Ikonomakis2005}. By comparing the classification methods, a voted classifier is ultimately constructed based on the documents and their features.

\section{Literature Review}
There are several works in the recent past on text classification. Li et al.\ \citep{Li2003} proposed a text classification technique using positive and unlabeled data. This paper defined two classes: one labeled as positive and another as unlabeled. The unlabeled data contained both positive and negative examples. The task was to identify the labeled data within the unlabeled class. The key finding was that only one class needed explicit labels, while the other did not.

Aggarwal et al.\ \citep{Aggarwal2012} surveyed various approaches to text categorization. This work discussed multiple aspects of automatic text categorization and compared techniques along with their benefits and shortcomings for different applications.

Tong et al.\ \citep{Tong2001} applied support vector machines (SVMs) to text categorization. Their study demonstrated that many relevant features could be discovered when applying SVMs. In contrast, our work employs an extensive package of SVM classification techniques optimized for text classification, with a voting process to ensure the best results.

Liu et al.\ \citep{Liu2004} discussed classifying text by labeling words instead of documents. While this method can be effective, it requires documents to be enriched with many relevant words for each class. In our work, we instead use Named Entity Recognition (NER) in Natural Language Processing (NLP) to simplify identifying relevant features for document labeling.

Sch{\"u}tze \citep{Schutze2008} estimated the conditional probability of a word given a class as its relative frequency in documents belonging to that class. However, Bernoulli Na{\"i}ve Bayes can be inefficient for long documents, as it does not account for multiple occurrences of words.

Tang et al.\ \citep{Tang2016} employed a Bayesian approach for automated text categorization, using information gain (IG) and maximum discrimination (MD) for feature selection. In our work, we use the document frequency metric for feature selection and apply Na{\"i}ve Bayes to compare results with previous works \citep{Ikonomakis2005}.

\section{Methodologies}

Document frequency (DF) is used for feature selection in text classification \citep{Ikonomakis2005}. It is defined as
\begin{equation}
  DF(t_k) = P(t_k), \label{eq:df}
\end{equation}
where $DF(t_k)$ is the document frequency of term $t_k$ and $P(t_k)$ is its probability of occurrence. The classifier models used for text classification are described below.

\subsection{Na\"ive Bayes Model}
The Na\"ive Bayes model includes Multinomial Na\"ive Bayes and Bernoulli Na\"ive Bayes \citep{McCallum1998}.

\subsubsection{Multinomial Na\"ive Bayes}
Multinomial Na\"ive Bayes models the count of terms in documents \citep{McCallum1998}. This variation accounts for the number of occurrences of a term $t$ in training documents from class $C$, including multiple occurrences.

\begin{algorithm}[H]
\caption{Multinomial Na\"ive Bayes Training}
\label{alg:mnb-train}
\begin{algorithmic}[1]
\State $V \gets \textsc{ExtractVocabulary}(D)$
\State $N \gets \textsc{CountDocs}(D)$
\For{each $c \in C$}
  \State $N_c \gets \textsc{CountDocsInClass}(D,c)$
  \State $\textit{prior}[c] \gets N_c/N$
  \State $\textit{text}_c \gets \textsc{ConcatTextOfAllDocsInClass}(D,c)$
  \For{each $t \in V$}
    \State $T_{ct} \gets \textsc{CountTokensOfTerm}(\textit{text}_c, t)$
  \EndFor
  \For{each $t \in V$}
    \State $\textit{condprob}[t][c] \gets \dfrac{T_{ct}+1}{\sum_{t'}(T_{ct'}+1)}$
  \EndFor
\EndFor
\State \Return $V,\ \textit{prior},\ \textit{condprob}$
\end{algorithmic}
\end{algorithm}

\begin{algorithm}[H]
\caption{Multinomial Na\"ive Bayes Application}
\label{alg:mnb-apply}
\begin{algorithmic}[1]
\State $W \gets \textsc{ExtractTokensFromDoc}(V,d)$
\For{each $c \in C$}
  \State $\textit{score}[c] \gets \log \textit{prior}[c]$
  \For{each $t \in W$}
    \State $\textit{score}[c] \gets \textit{score}[c] + \log \textit{condprob}[t][c]$
  \EndFor
\EndFor
\State \Return $\operatorname*{arg\,max}\limits_{c \in C}\ \textit{score}[c]$
\end{algorithmic}
\end{algorithm}

\subsubsection{Bernoulli Na\"ive Bayes}
This variant generates each token from the vocabulary and assigns $1$ if the token appears in the document and $0$ otherwise \citep{McCallum1998}.

\begin{algorithm}[H]
\caption{Bernoulli Na\"ive Bayes Training}
\label{alg:bnb-train}
\begin{algorithmic}[1]
\State $V \gets \textsc{ExtractVocabulary}(D)$
\State $N \gets \textsc{CountDocs}(D)$
\For{each $c \in C$}
  \State $N_c \gets \textsc{CountDocsInClass}(D,c)$
  \State $\textit{prior}[c] \gets N_c/N$
  \For{each $t \in V$}
    \State $N_{ct} \gets \textsc{CountDocsInClassContainingTerm}(D,c,t)$
    \State $\textit{condprob}[t][c] \gets \dfrac{N_{ct}+1}{N_c+2}$
  \EndFor
\EndFor
\State \Return $V,\ \textit{prior},\ \textit{condprob}$
\end{algorithmic}
\end{algorithm}

\begin{algorithm}[H]
\caption{Bernoulli Na\"ive Bayes Application}
\label{alg:bnb-apply}
\begin{algorithmic}[1]
\State $V_d \gets \textsc{ExtractTermsFromDoc}(V,d)$
\For{each $c \in C$}
  \State $\textit{score}[c] \gets \log \textit{prior}[c]$
  \For{each $t \in V$}
    \If{$t \in V_d$}
      \State $\textit{score}[c] \gets \textit{score}[c] + \log \textit{condprob}[t][c]$
    \Else
      \State $\textit{score}[c] \gets \textit{score}[c] + \log\!\big(1 - \textit{condprob}[t][c]\big)$
    \EndIf
  \EndFor
\EndFor
\State \Return $\operatorname*{arg\,max}\limits_{c \in C}\ \textit{score}[c]$
\end{algorithmic}
\end{algorithm}

\subsection{Linear Classifier Model}

The linear classifier model is based on gradient-descent methods, including Logistic Regression and Stochastic Gradient Descent (SGD).

\subsubsection{Logistic Regression }
The probability of assigning document $d^{(i)}$ to class $c=j$ is \citep{Genkin2007}
\begin{equation}
P\!\left(c^{(i)}=j \mid d^{(i)};\theta\right)
= \frac{\exp(\theta_j^\top d^{(i)})}{\sum_{l=1}^{k} \exp(\theta_l^\top d^{(i)})},
\end{equation}
where $\theta_1,\ldots,\theta_k \in \mathbb{R}^{n+1}$ are the parameters. The hypothesis vector is
\begin{equation}
h_\theta(d^{(i)}) =
\begin{bmatrix}
P(c^{(i)}{=}1 \mid d^{(i)};\theta)\\
\vdots\\
P(c^{(i)}{=}k \mid d^{(i)};\theta)
\end{bmatrix}
=
\frac{1}{\sum_{j=1}^{k}\exp(\theta_j^\top d^{(i)})}
\begin{bmatrix}
\exp(\theta_1^\top d^{(i)})\\
\vdots\\
\exp(\theta_k^\top d^{(i)})
\end{bmatrix}.
\end{equation}
The (multiclass) cross-entropy cost is
\begin{equation}
J(\theta) = -\frac{1}{m} \sum_{i=1}^{m} \sum_{j=1}^{k}
\mathbf{1}\{c^{(i)}=j\}\,
\log \frac{\exp(\theta_j^\top d^{(i)})}{\sum_{l=1}^{k}\exp(\theta_l^\top d^{(i)})},
\end{equation}
with gradient
\begin{equation}
\nabla_{\theta_j} J(\theta) = -\frac{1}{m}\sum_{i=1}^{m}
\big[\; d^{(i)} \big(\mathbf{1}\{c^{(i)}=j\} - P(c^{(i)}=j \mid d^{(i)};\theta)\big)\big].
\end{equation}

\subsubsection{Stochastic Gradient Descent }
For a training pair $(d^{(i)}, c^{(i)})$, a simple squared-error cost is \citep{Bottou2010}
\begin{equation}
\mathrm{cost}\!\left(\theta;(d^{(i)},c^{(i)})\right)
= \frac{1}{2}\big(h_\theta(d^{(i)})-c^{(i)}\big)^2,
\end{equation}
and the empirical objective is
\begin{equation}
J_{\text{train}}(\theta) = \frac{1}{2m}\sum_{i=1}^{m}
\mathrm{cost}\!\left(\theta;(d^{(i)},c^{(i)})\right).
\end{equation}

\begin{algorithm}[H]
\caption{Stochastic Gradient Descent (one epoch)}
\begin{algorithmic}[1]
\State Randomly shuffle the dataset
\For{$i = 1$ to $m$} \Comment{training examples}
  \For{$j = 0$ to $n$} \Comment{features}
    \State $\theta_j \gets \theta_j - \alpha \,\big(h_\theta(d^{(i)}) - c^{(i)}\big)\, d^{(i)}_j$
  \EndFor
\EndFor
\end{algorithmic}
\end{algorithm}

Here, $\alpha$ is the learning rate (often decayed over iterations) to promote convergence of $\theta$.

\subsection{Support Vector Machine Model}

The Support Vector Machine (SVM) model includes Support Vector Clustering (SVC) and Linear Support Vector Clustering (Linear SVC) \citep{Cortes1995}.  
The algorithms are applied using an optimized SVM library, specifying parameters and kernel functions to measure similarity between classes and documents.

\subsubsection{Support Vector Clustering (SVC)}
In SVC \citep{Cortes1995}, the Gaussian kernel is defined as
\begin{equation}
f_G = -\frac{\| D - l^{(i)} \|}{\sigma^2},
\end{equation}
where $D$ is the set of documents, $l^{(i)}$ is a landmark equal to $d^{(i)}$, and $\sigma$ is the kernel bandwidth.  
Applying this kernel is particularly useful when the number of features $n$ is small and the number of training examples $m$ is large.

\subsubsection{Linear Support Vector Clustering (Linear SVC)}
To predict whether a document $d$ belongs to class $c$, Linear SVC evaluates the decision boundary
\begin{equation}
\theta^\top d \geq 0.
\end{equation}
If the inequality holds, then $d$ is assigned to class $c$; otherwise, it is not.  
Applying a linear kernel is especially effective when the number of features $n$ is high and the number of training examples $m$ is small.

\subsection{Artificial Neural Network Model}
The Artificial Neural Network (ANN) model includes only the Backpropagation Network. The documents are given as inputs to the input layer. There are two hidden layers that adapt the weights and thresholds. The outputs correspond to the documents associated with their proper classes. 

The advantage of this model is that the number of documents and classes can vary, which allows adjusting the number of input and output nodes. Sigmoid and Gaussian functions are used as transfer functions in the Backpropagation Network.

\begin{figure}[htbp]
  \centering
  \begin{tikzpicture}[
    neuron/.style={circle, draw=black!80, fill=#1!30, minimum size=16pt},
    input neuron/.style={neuron=blue},
    hidden neuron/.style={neuron=orange},
    output neuron/.style={neuron=green},
    >=stealth,
    shorten >=1pt,
    node distance=1cm,
  ]
    \def\nInput{3}
    \def\nHiddenA{5}
    \def\nHiddenB{5}
    \def\nOutput{3}

    \foreach \i in {1,...,\nInput} {
      \node[input neuron] (I-\i) at (0,-\i) {$x_{\i}$};
    }

    \foreach \i in {1,...,\nHiddenA} {
      \node[hidden neuron] (H1-\i) at (2,-\i-0.5) {$a^{(2)}_{\i}$};
    }

    \foreach \i in {1,...,\nHiddenB} {
      \node[hidden neuron] (H2-\i) at (4,-\i-0.5) {$a^{(3)}_{\i}$};
    }

    \foreach \i in {1,...,\nOutput} {
      \node[output neuron] (O-\i) at (6,-\i-1) {$y_{\i}$};
    }

    \foreach \i in {1,...,\nInput}
      \foreach \j in {1,...,\nHiddenA}
        \draw[->] (I-\i) -- (H1-\j);

    \foreach \i in {1,...,\nHiddenA}
      \foreach \j in {1,...,\nHiddenB}
        \draw[->] (H1-\i) -- (H2-\j);

    \foreach \i in {1,...,\nHiddenB}
      \foreach \j in {1,...,\nOutput}
        \draw[->] (H2-\i) -- (O-\j);
  \end{tikzpicture}
  \caption{%
    Fully connected feedforward artificial neural network with two hidden layers. 
    The input layer has 3 neurons ($x_1,x_2,x_3$), each connected to 5 neurons in the first hidden layer ($a^{(2)}_1,\dots,a^{(2)}_5$), 
    which in turn connect to 5 neurons in the second hidden layer ($a^{(3)}_1,\dots,a^{(3)}_5$), 
    and finally to 3 output neurons ($y_1,y_2,y_3$). 
    Every neuron in each layer connects to every neuron in the next, illustrating the dense topology of a multilayer perceptron.%
  }
  \label{fig:ann-architecture}
\end{figure}
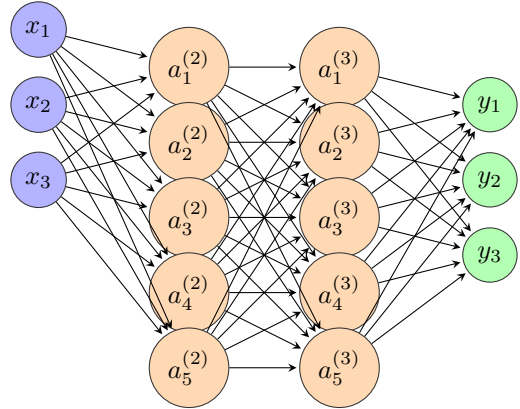

\subsubsection{Backpropagation Network}
The Backpropagation algorithm is described as follows \citep{Rumelhart1986}:

\begin{algorithm}[H]
\caption{Backpropagation Algorithm}
\begin{algorithmic}[1]
\State Initialize the training set: 
\[
\{(d^{(1)}, c^{(1)}), (d^{(2)}, c^{(2)}), \ldots, (d^{(m)}, c^{(m)})\}
\]
\State Set $\Delta_{ij}^{(l)} := 0$ for all layers $l$, nodes $i$ and $j$.
\For{$i \gets 1$ to $m$}
  \State Set $a^{(1)} = d^{(i)}$.
  \State Perform forward propagation to compute $a^{(l)}$ for $l = 1,2,\ldots,L$.
  \State Using $c^{(i)}$, compute $\delta^{L} = a^L - c^{(i)}$.
  \State Compute $\delta^{(L-1)}, \delta^{(L-2)}, \ldots, \delta^{(2)}$.
  \State Update: $\Delta_{ij}^{(l)} := \Delta_{ij}^{(l)} + a_j^{(l)} \delta_i^{(l+1)}$.
\EndFor
\end{algorithmic}
\end{algorithm}

Here, $m$ is the number of training examples, $L$ is the total number of layers, $\delta_j^{(l)}$ is the error of node $j$ in layer $l$, and $a_j^{(l)}$ is the activation value of node $j$ in layer $l$.

The weight update term is defined as:
\[
D_{ij}^{(l)} :=
\begin{cases}
\frac{1}{m}\,\Delta_{ij}^{(l)} + \lambda\,\Theta_{ij}^{(l)}, & j \neq 0,\\[6pt]
\frac{1}{m}\,\Delta_{ij}^{(l)}, & j = 0.
\end{cases}
\]

The cost function is given by:
\begin{equation}
J(\Theta) = -\frac{1}{m} \left[ \sum_{i=1}^{m} \Big( c^{(i)} \log(h_\Theta(d^{(i)})) + (1 - c^{(i)}) \log(1 - h_\Theta(d^{(i)})) \Big) \right] 
+ \frac{\lambda}{2m} \sum_{l=1}^{L-1} \sum_{i=1}^{S_l} \sum_{j=1}^{S_{l+1}} (\Theta_{ji}^{(l)})^2
\end{equation}

The gradient of the cost function is:
\begin{equation}
\frac{\partial}{\partial \Theta_{ij}^{(l)}} J(\Theta) = D_{ij}^{(l)}
\end{equation}

\section{Results}

The metric of accuracy depends on four features \citep{Ikonomakis2005}:
\begin{itemize}
    \item \textbf{True positives (TP)}: relevant documents correctly identified as the proper class.
    \item \textbf{True negatives (TN)}: irrelevant documents correctly identified as an improper class.
    \item \textbf{False positives (FP)}: irrelevant documents incorrectly identified as proper class (type I error).
    \item \textbf{False negatives (FN)}: relevant documents incorrectly identified as improper class (type II error).
\end{itemize}

From these, two parameters are used to calculate the F1 score.

\subsection*{Precision}
\begin{equation}
\text{Precision} = \frac{\text{TP}}{\text{TP} + \text{FP}}
\end{equation}

\subsection*{Recall}
\begin{equation}
\text{Recall} = \frac{\text{TP}}{\text{TP} + \text{FN}}
\end{equation}

\subsection*{F1 Score}
\begin{equation}
F1 = \frac{2 \times \text{Precision} \times \text{Recall}}{\text{Precision} + \text{Recall}}
\end{equation}

\noindent Using the F1 score metric, the accuracy percentage of text classifiers on different datasets is summarized in Table~\ref{tab:results}.

\begin{table}[H]
\centering
\caption{Accuracy (\%) of Text Classification by F1 Score}
\label{tab:results}
\begin{tabular}{@{}lccc@{}}
\toprule
\textbf{Classifier Model} & \textbf{Reuters Corpus \citep{ReutersCorpus}} & \textbf{Brown Corpus \citep{BrownCorpus}} & \textbf{Movie Review Corpus \citep{CornellReviews}} \\ \midrule
\multicolumn{4}{c}{\textbf{Naïve Bayes Model}} \\ \midrule
Multinomial Naïve Bayes   & 72.0 & 72.5 & 76.0 \\
Bernoulli Naïve Bayes     & 75.0 & 78.0 & 79.0 \\ \midrule
\multicolumn{4}{c}{\textbf{Linear Classifier Model}} \\ \midrule
Logistic Regression       & 73.5 & 79.5 & 74.5 \\
Stochastic Gradient Descent & 76.0 & 83.5 & 81.5 \\ \midrule
\multicolumn{4}{c}{\textbf{SVM Model}} \\ \midrule
SVC                       & 78.0 & 78.0 & 79.5 \\
Linear SVC                & 83.0 & 77.0 & 80.5 \\ \midrule
\multicolumn{4}{c}{\textbf{Artificial Neural Network Model}} \\ \midrule
Back Propagation Network  & 89.0 & 93.0 & 94.5 \\ 
\textbf{Voted Classifier} & \textbf{89.0} & \textbf{93.0} & \textbf{94.5} \\ \bottomrule
\end{tabular}
\end{table}

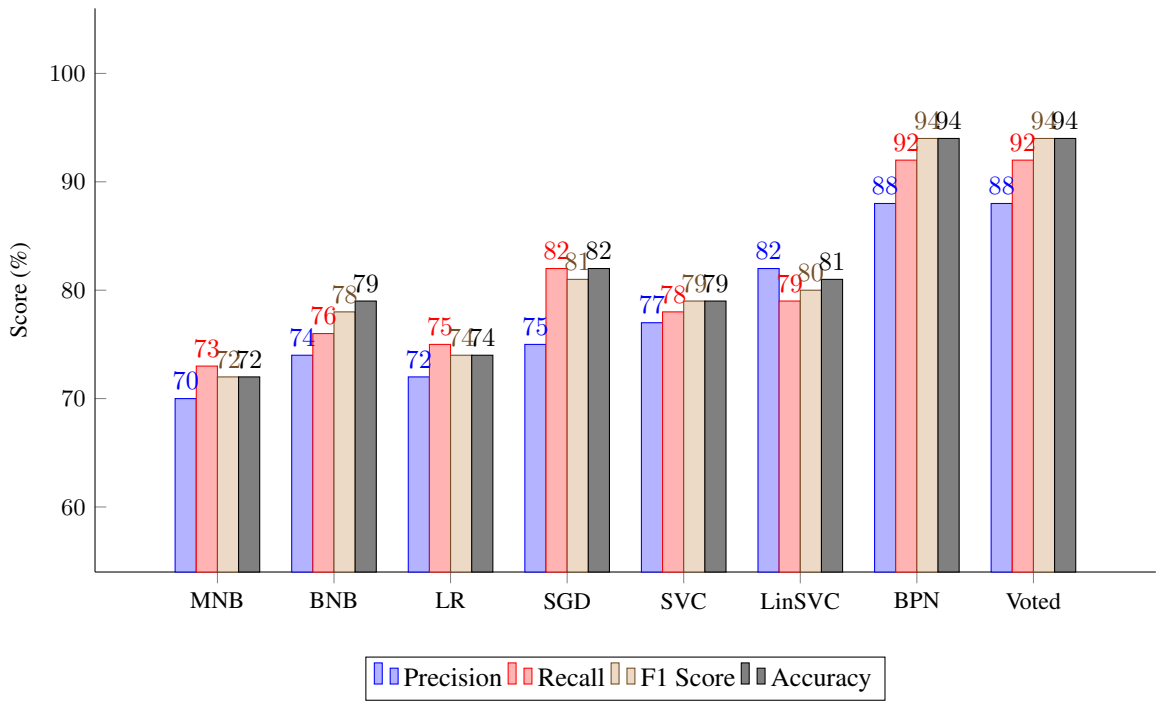
\begin{figure}[H]
\centering
\begin{tikzpicture}
\begin{axis}[
    ybar=0pt,
    bar width=8pt,
    width=0.95\textwidth,
    height=0.55\textwidth,
    enlargelimits=0.15,
    legend style={at={(0.5,-0.15)}, anchor=north, legend columns=-1},
    ylabel={Score (\%)},
    symbolic x coords={
        MNB, BNB, LR, SGD, SVC, LinSVC, BPN, Voted
    },
    xtick=data,
    nodes near coords,
    nodes near coords align={vertical},
    ymin=60, ymax=100,
    axis x line*=bottom,
    axis y line*=left,
    tick label style={font=\small},
    label style={font=\small},
    legend cell align={left}
]

\addplot coordinates {
    (MNB,70) (BNB,74) (LR,72) (SGD,75) (SVC,77) (LinSVC,82) (BPN,88) (Voted,88)
};

\addplot coordinates {
    (MNB,73) (BNB,76) (LR,75) (SGD,82) (SVC,78) (LinSVC,79) (BPN,92) (Voted,92)
};

\addplot coordinates {
    (MNB,72) (BNB,78) (LR,74) (SGD,81) (SVC,79) (LinSVC,80) (BPN,94) (Voted,94)
};

\addplot coordinates {
    (MNB,72) (BNB,79) (LR,74) (SGD,82) (SVC,79) (LinSVC,81) (BPN,94) (Voted,94)
};

\legend{Precision, Recall, F1 Score, Accuracy}
\end{axis}
\end{tikzpicture}
\caption{Performance comparison of classifiers on text classification tasks (Precision, Recall, F1 Score, and Accuracy).}
\label{fig:results}
\end{figure}

\section{Conclusion}
This research aimed to identify the best classification method and to select the voted classifier, i.e., the classifier that achieves the highest accuracy. It also demonstrates that the performance of text document classification depends not only on the classification method but also on how well the associated corpus is organized. 

Each corpus dataset was divided into three categories: 60\% of the data was used for training, 20\% for cross-validation, and the remaining 20\% for testing. From the results, it can be observed that classification performance may vary slightly across time and datasets, but the accuracy approximation remains consistent. 

It was also found that the Artificial Neural Network (ANN) achieved higher accuracy compared to the other classifier models. This is because, until the classification process becomes satisfactory, the ANN continues to iterate and optimize. Although it requires more iterations and execution time, it provides the most accurate classification of text documents. 

This research combines methods from Machine Learning and fundamental Natural Language Processing techniques. As text documents continue to grow rapidly worldwide, processing and categorizing them into proper classes will remain a major concern in future research.

\section*{Acknowledgment}
The authors would like to thank Biprodip Pal, Assistant Professor, Department of Computer Science and Engineering, Rajshahi University of Engineering and Technology, for his support and enthusiasm. The authors are also grateful to Professor Andrew Ng of Stanford University and his online course on Machine Learning in Coursera \citep{CourseraML}.

\end{document}